\pdfoutput=1

\documentclass[11pt]{article}

\usepackage[final]{acl}

\usepackage{times}
\usepackage{latexsym}

\usepackage[T1]{fontenc}

\usepackage[utf8]{inputenc}

\usepackage{microtype}

\usepackage{inconsolata}

\usepackage{graphicx}
\usepackage{amsmath}
\usepackage{amssymb}
\usepackage{multirow}
\usepackage{algorithm, algorithmic}
\usepackage[most]{tcolorbox}
\tcbuselibrary{breakable}
\usepackage{color}
 
\definecolor{myblue}{rgb}{0,0,1}

\title{PSC: Extending Context Window of Large Language Models via Phase Shift Calibration}

\author{Wenqiao Zhu \and Chao Xu \and Lulu Wang \and  Jun Wu \\
    HiThink Research \\
    \texttt{\{zhuwenqiao,xuchao3,wanglulu2,wujun2\}@myhexin.com} \\
}

\begin{document}
\maketitle
\begin{abstract}
Rotary Position Embedding (RoPE) is an efficient position encoding approach and is widely utilized in numerous large language models (LLMs). Recently, a lot of methods have been put forward to further expand the context window based on RoPE. The core concept of those methods is to predefine or search for a set of factors to rescale the base frequencies of RoPE. Nevertheless, it is quite a challenge for existing methods to predefine an optimal factor due to the exponential search space. In view of this, we introduce PSC (Phase Shift Calibration), a small module for calibrating the frequencies predefined by existing methods. With the employment of PSC, we demonstrate that many existing methods can be further enhanced, like PI, YaRN, and LongRoPE. We conducted extensive experiments across multiple models and tasks. The results demonstrate that (1) when PSC is enabled, the comparative reductions in perplexity increase as the context window size is varied from 16k, to 32k, and up to 64k. (2) Our approach is broadly applicable and exhibits robustness across a variety of models and tasks.
The code can be found at https://github.com/WNQzhu/PSC.
\end{abstract}

\section{Introduction}
Large-scale language models (LLMs) have shown impressive results across a variety of natural language processing (NLP) applications. For instance, OpenAI has shown that GPT-4 \citep{OpenAI-GPT4} can perform at a level comparable to humans in a range of professional tasks. Additionally, open-source models such as LLaMA2 \citep{Touvron2023Llama2O} and Mistral \citep{Jiang2023Mistral7} have made significant contributions to the advancement and practical application of LLMs in both research and industry.
However, one significant challenge that LLMs face is handling tasks that require processing long context, such as responding to questions based on multiple documents and summarizing lengthy texts such as books. In these scenarios, the perplexity of the responses can increase substantially, leading to a notable decrease in the performance of LLMs. Therefore, equipping LLMs with long-range ability has become a critical and pressing issue for both academic and commercial sectors.

An intuitive method is to fine-tune a pre-trained Transformer with a longer context length. Nevertheless, there are two limitations: first, models trained in this manner adapt to long context lengths very slowly \citep{Chen2023ExtendingCW}; second, fine-tuning updates all model parameters is memory-inefficient which prevents the model from adapting to a large context length \citep{longlora}.

Optimizing position encodings is another major direction for extending the context window of LLMs \citep{jin2024llm}. 
The original Transformer \citep{vaswani2017attention} that serves as the core component of LLMs uses sinusoidal functions of various frequencies to enhance the model's extrapolate capability. It could be regarded as an absolute position encoding mechanism.
Since then, relative positional encoding techniques such as RoPE \citep{su2021roformer} and ALiBi \citep{alibi2022} have further increased the length extrapolation of Transformers.
Despite the effectiveness, many existing pre-trained LLMs that use these positional encoding methods exhibit weak extrapolation capabilities.
For example, LLaMA \citep{Touvron2023LLaMAOA} with 2048 predefined context size explodes perplexity metric when the input texts length is larger than 4096 \citep{Chen2023ExtendingCW}.

Recently, new positional encoding schemes have been proposed to overcome such limitations.
\citep{Chen2023ExtendingCW} and \citep{kaiokendev} show that the effective context size could be extended
by modifying RoPE via Position Interpolation, which has a much smaller upper bound than the extrapolated method and is more stable \citep{Chen2023ExtendingCW}.
Neural Tangent Kernel (NTK) theory shows that it's difficult for multilayer perceptron (MLP) to learn high-frequency information in a low-dimensional domain. Therefore, NTK-based methods take the high-frequency information into account \citep{block97-ntk-aware, block97-ntk-by-part, emozilla-ntk}.
Furthermore, YaRN hypothesizes that previous methods lead to a closer embedding distribution and remedy the issue by using different interpolating schemes at different frequencies \citep{peng2023yarn}.
The shared characteristic of previous methods is that they utilize predefined frequency rescale factors.
Some algorithms leverage optimal methods to estimate optimal frequencies directly, such as LongRoPE \citep{longrope} and CLEX \citep{damonlpsg2023clex}.
However, due to the exponential search space complexity, it is challenging for those methods to estimate an optimal frequency; they also need heavy searching cost, for instance, it costs LongRoPE nearly 3 days to search an optimal frequency for a 256k context window using an A100 GPU.

While existing techniques for encoding positional information are adept at handling long-range dependencies, they often depend on fixed patterns or necessitate extensive searches within large parameter spaces. As a result, adapter-based approaches such as LoRA \citep{hu2022lora} have been utilized to further enhance performance. Nonetheless, these methods still face limitations, primarily due to the low rank of the adapter weights \citep{Biderman2024LoRALL} and the inherently high-rank nature of long-context tasks.

In this work, we introduce Phase Shift Calibration to assist position encoding methods to improve their long-range capabilities.
The main idea is that we propose a module to calibrate the predefined frequency to approximate the optimal frequency.
To this end, we first present that there is a rotary transformation between the actual frequencies and the optimal frequencies. The transformation can be represented as a block diagonal matrix. It is full-rank if the predefined frequencies are far from the optimal ones. Hence, it is challenging for low-rank adapter methods such as LoRA to
learn the transformation. To remedy this issue, we introduce a calibration module into the base model, which approximates the rotary transformation matrix and helps calibrate the predefined frequencies to the ideal position.
We conduct extensive experiments across different LLMs, position encoding schemes, and various long-context tasks. The results demonstrate the effectiveness of our methods.
\section{Preliminaries and Related Work}
\paragraph{Rotary Position Embedding (RoPE).}
Transformer models leverage positional information to exploit the order of tokens within texts.
In our research, we concentrate on Rotary Position Embedding (RoPE) \citep{su2021roformer} and its derivatives.
RoPE acts as the positional encoding technique used across various Large Language Models (LLMs), such as the LLaMA \citep{Touvron2023LLaMAOA} and the Mistral model \citep{Jiang2023Mistral7}.
Given a sequence of N word embeddings $\{\mathbf{x}_i\}_{i=1}^{\mathcal{N}}$, where $\mathbf{x}_i$ is a $d$-dimensional vector and $d$ is the dimension of the embedding.
RoPE applies a rotary transformation to each query/key embedding in a pairwise manner.
Take $d=2$ for example, RoPE converts each vector into the query vector and key vector via a transformation in a complex space: 
\begin{equation}
  \mathbf{q}_m = f_q(\mathbf{x}_m, m) = (\mathbf{W}_q \mathbf{x}_m) e ^{i m \theta} \label{eq:2d:q}
\end{equation}
\begin{equation}
  \mathbf{k}_n = f_k(\mathbf{x}_n, n) = (\mathbf{W}_k \mathbf{x}_n) e ^{i n \theta} \label{eq:2d:k}
\end{equation}
where $m, n$ are the position index, $i \doteq \sqrt{-1}$ is the imaginary unit.
After rotary transformation, the attention scores are calculated as
\begin{equation}
\textit{softmax}\left(\frac{\mathbf{q}_m^T \mathbf{k}_n}{\sqrt{d}}\right)
\end{equation}
The rotary transformation introduces an $m-n$ term in the attention score:
\begin{align}
  \mathbf{q}_m^T \mathbf{k}_n & = Re \langle f_q(\mathbf{x}_m, m),  f_k(\mathbf{x}_n, n) \rangle \notag \\
  &= \left(\mathbf{W}_q \mathbf{x}_m\right)^T \left( \begin{array}{ll} c_1 & -c_2 \\ c_2 & c_1 \\\end{array} \right) \mathbf{W}_k \mathbf{x}_n \notag \\
  &\doteq g(\mathbf{x}_m, \mathbf{x}_n, m - n),
\end{align}
where
\begin{equation}
c_1=cos (m-n)\theta, \quad c_2=sin (m-n)\theta. \notag
\end{equation}
Hence, RoPE possesses the capability of encoding relative positional information via absolute positional encoding.
For a general form where $d \ge 2$,  RoPE divides the $d$-dimensional space into $d/2$ 2D complex sub-spaces:
\begin{align}
& (\mathbf{x}_m)_1, (\mathbf{x}_m)_2, \cdots, (\mathbf{x}_m)_d \mapsto  \notag \\ &  (\mathbf{x}_m)_1 +  i (\mathbf{x}_m)_2, \cdots, (\mathbf{x}_m)_{d-1} + i(\mathbf{x}_m)_d
\end{align}
In matrix form, the rotary-transformed query and key can be expressed as:
\begin{equation}
 f_{q} = \mathbf{R}^d_{\Theta, m} \mathbf{W}_{q} \mathbf{x}_m 
\end{equation}
\begin{equation}
 f_{k} = \mathbf{R}^d_{\Theta, n} \mathbf{W}_{k} \mathbf{x}_n 
\end{equation}
where
\begin{equation}
  \mathbf{R}^d_{\Theta, m} = \left(
  \begin{array}{llll}
    \mathbf{B}_{m,1} & \mathbf{0} & \mathbf{0} & \mathbf{0} \\
    \mathbf{0}       & \mathbf{B}_{m,2} & \mathbf{0} & \mathbf{0} \\
    \mathbf{0}       & \mathbf{0}      &  \cdots    & \mathbf{0} \\
    \mathbf{0}       & \mathbf{0}      & \mathbf{0} & \mathbf{B}_{m,d/2} \\
  \end{array}
  \right), \notag
\end{equation}
\begin{equation}
\mathbf{B}_{m,i} = \left(
  \begin{array}{ll}
    \cos m\theta_i & -\sin m\theta_i  \\
    \cos m\theta_i & \sin m \theta_i  \\
  \end{array}
\right) \notag
\end{equation}
and $\Theta = \{\theta_i = b ^{-2(i-1)/d}, i\in [1,2,\cdots, d/2]\}$ is the predefined frequencies.
In many models, $b$ is set to $10^4$.

\begin{figure}[t]
\begin{center}
  \includegraphics[width=\linewidth]{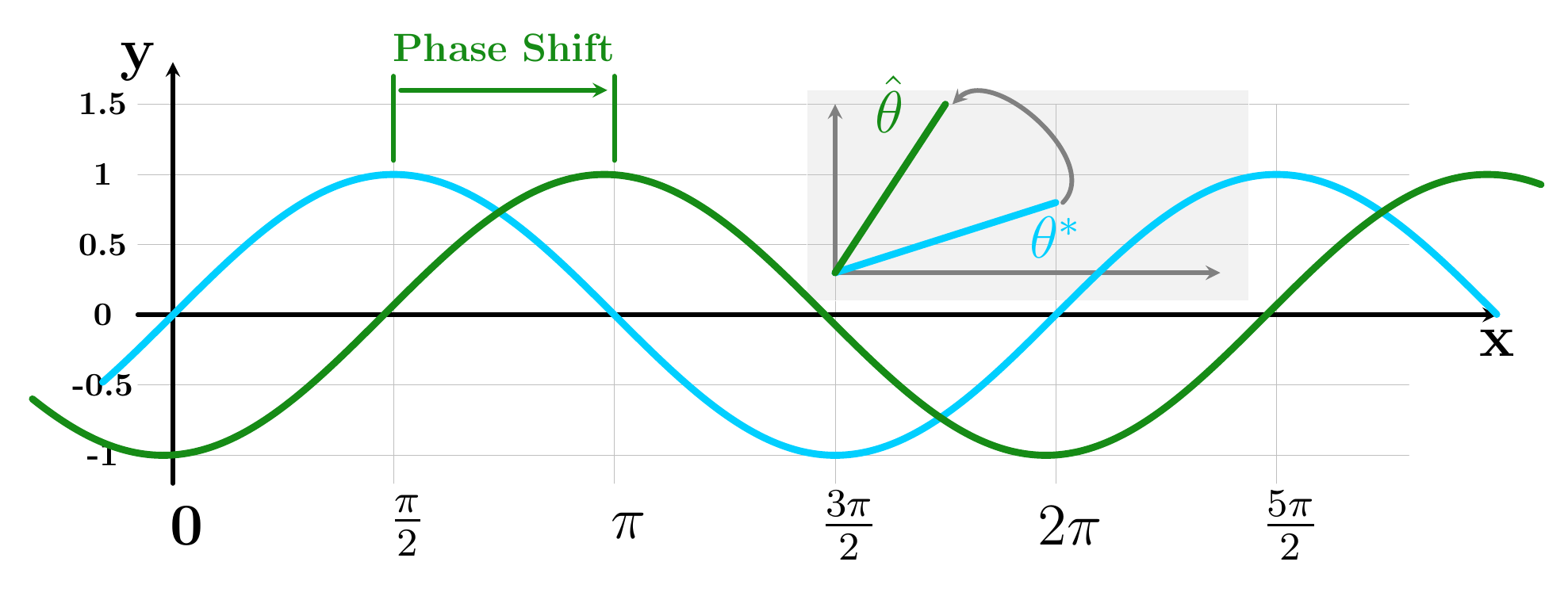}
\end{center}
\caption{Phase shift leads to the $\sin/\cos$ values deviating from their optimal positions. The $\theta^*$ is assumed to be an optimal frequency.}
\label{shift_tikz}
\end{figure}
\paragraph{RoPE Extensions.}
Various RoPE-like positional encoding schemes have been proposed to enhance the capabilities of long-range dependencies.
We can unify them into the following general form:
\begin{equation}
  f_q := f_q(\mathbf{x}_m, m, h(\theta_i))
\end{equation}
Position Interpolation \citep{Chen2023ExtendingCW} originally proposed to interpolate the position index $m$
by modifing it into $\frac{L}{L^\prime}m$, where $L$ is the predefined context size and $L^\prime$ is the new context window beyond the
pre-trained limit. Hence, $h^{\textit{PI}}(\theta_i) = \frac{L}{L^\prime} \theta_i$.
The NTK-aware scheme modifies RoPE by taking into account the loss of high-frequency components through the utilization of the following formulation: $h^{\textit{NTK}}(\theta_i) = \left(b \cdot s ^{\frac{d}{d-2}}\right)^{-2i/d}$, where $s$ is the scaling factor.
YaRN \citep{peng2023yarn} employs extrapolations in the high-frequency domain, interpolations in the low-frequency domain, and a blend of both in the intermediate frequencies. The frequency function $h^{\textit{YaRN}}(\theta_i) = (1-\gamma) \frac{\theta_i}{s} + \gamma \theta_i$, where $\gamma$ is the blend factor.
LongRoPE \citep{longrope} utilizes evoluation-based search to
estimate optimal scale factors $s_o$, and the actual frequencies are scaled to $h^{\textit{LongRoPE}}(\theta_i) = \frac{\theta_i}{s_o}$.
\begin{figure*}[t]
\begin{center}
  \includegraphics[width=\linewidth]{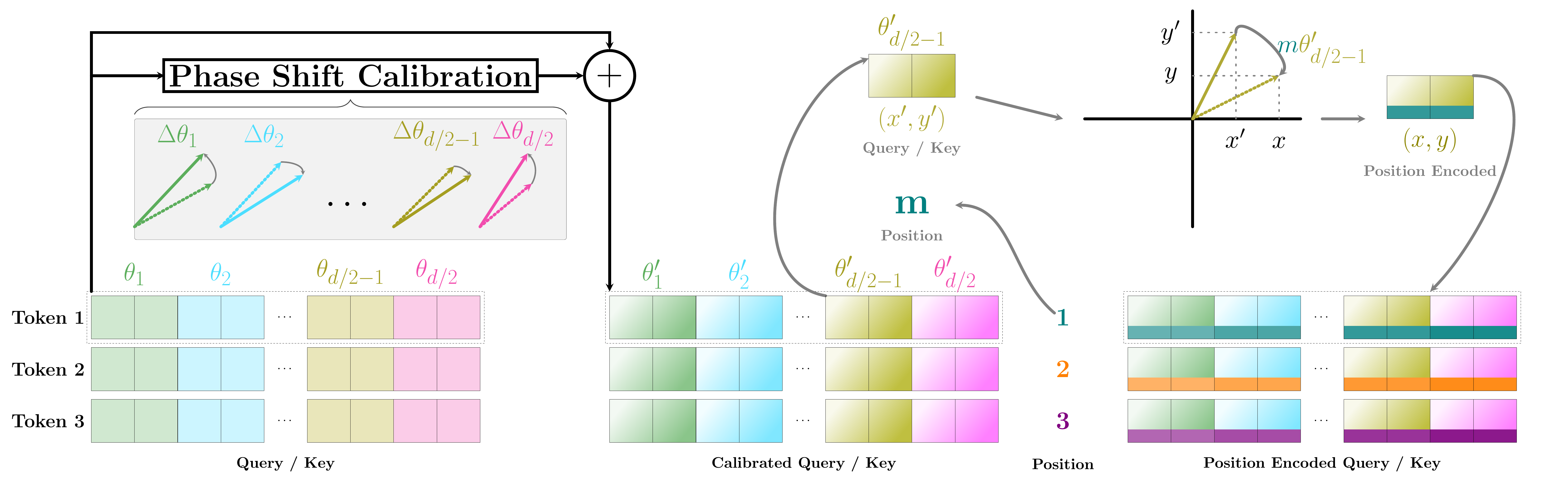}
\end{center}
\caption{The embeddings are calibrated to an ideal position, and then existing position encode methods are adopted.}
\label{framework}
\end{figure*}
\paragraph{Low-rank Adaption.}
LoRA \citep{hu2022lora} posits that the weight adjustments in pre-trained models are characterized by a low intrinsic rank during adaptation.
Given a pre-trained weight matrix $\mathbf{W} \in R^{d\times k}$, it is updated with a low-rank decomposition $\mathbf{W}+\Delta \mathbf{W} = \mathbf{W} + \mathbf{B} \mathbf{A}$, where $\mathbf{B} \in R^{d\times r}$,
$\mathbf{A} \in R^{r\times k}$, and $r \ll \min(d,k)$.
During training, $\mathbf{W}$ remains fixed, while $\mathbf{A}$ and $\mathbf{B}$ are trainable.

\section{Methodology}
\subsection{Phase Shift}

Let $\theta^*$ denote the optimal frequency for long context extension of a large language model,
$\hat{\theta}$ the frequency predefined or estimated by some algorithms, such as PI or LongRoPE.
It is challenging to predefine a frequency $\hat{\theta}$ that is exactly equal to $\theta^*$ due to the exponential search space. The suboptimal frequencies cause the $\sin/\cos$ values to move out of the ideal position, as shown in Figure \ref{shift_tikz}. As a result,  there exists a rotary transformation between the ideal position encoded embeddings and the actual embeddings:
\begin{align}
  f^*_q(\mathbf{x}_m, m) &= (\mathbf{W}_q \mathbf{x}_m) e ^{im\theta^*} \notag \\
  & = (\mathbf{W}_q \mathbf{x}_m) e ^{im\theta^* + im\hat{\theta} - im\hat{\theta}} \notag \\
  & = \hat{f}_q(\mathbf{x}_m, m) e ^ {im(\theta^* - \hat{\theta})},
  \label{eq:ideal:q}
\end{align}
\begin{equation}
  f^*_k(\mathbf{x}_n, n) =  \hat{f}_k(\mathbf{x}_n, n) e ^ {in(\theta^* - \hat{\theta})},
\end{equation}
where $f^*_q(\mathbf{x}_m, m)$ and $f^*_k(\mathbf{x}_n, n)$ are the ideal query and key with the optimal frequencies;
$\hat{f}_q(\mathbf{x}_m, m)$ and $\hat{f}_k(\mathbf{x}_n, n)$ are the actual query and key with predefined frequenices.

In general form, the position-encoded query and key can be expressed as:
\begin{align}
  f^*_q(\mathbf{x}_m, m) & = \mathbf{\widetilde{R}}^d_{\Theta^* - \hat{\Theta}, m}\mathbf{R}^d_{\hat{\Theta}, m} \mathbf{W}_{q} \mathbf{x}_m \notag \\ & = \mathbf{\widetilde{R}}^d_{\Theta^* - \hat{\Theta}, m} \hat{f}_q(\mathbf{x}_m, m),
\end{align}
\begin{equation}
f^*_k(\mathbf{x}_n, n) = \mathbf{\widetilde{R}}^d_{\Theta^* - \hat{\Theta}, n} \hat{f}_q(\mathbf{x}_n, n), 
\end{equation}
where $\mathbf{\widetilde{R}}^d_{\Theta^* - \hat{\Theta}, n}$ is a block diagonal matrix with each block as
\[
\begin{bmatrix}
  \cos n(\theta^*_i - \hat{\theta}_i) & -\sin n(\theta^*_i - \hat{\theta}_i) \\
  \sin n(\theta^*_i - \hat{\theta}_i) &  \cos n(\theta^*_i - \hat{\theta}_i) \\ 
\end{bmatrix},
\]
$\Theta^*$ denotes the optimal frequency set $\{\theta^*_i\}$, $\hat{\Theta}$ denotes actual frequency set $\{\hat{\theta}_i\}$, and $i \in [0, d/2]$.
Let $\mathbf{\widetilde{W}} = \mathbf{R}^d_{\hat{\Theta}, m} \mathbf{W}_{q}$ and $\mathbf{I}$ denote the identity matrix, then $f^*_q = \mathbf{\widetilde{R}} \mathbf{\widetilde{W}} = \mathbf{\widetilde{W}} + (\mathbf{\widetilde{R}} - \mathbf{I})\mathbf{\widetilde{W}}$.
When low-rank adapter methods such as LoRA are employed to finetune the model, 
we need to utilize two low-rank matrices $\mathbf{A}$ and $\mathbf{B}$ to approximate the additional matrix.
Specifically, $\mathbf{B}\mathbf{A}_\text{LoRA} \rightarrow (\mathbf{\widetilde{R}} - \mathbf{I})\mathbf{\widetilde{W}}$.

Approximating the matrix becomes difficult if the pre-established frequencies are not ideal.
For instance, if none of the pre-established frequencies are optimal, then $\mathbf{\widetilde{R}} - \mathbf{I}$
becomes a matrix of full rank since it is a block diagonal matrix with all non-zero elements, while $\mathbf{B}\mathbf{A}$ remains a low-rank matrix.
The accuracy of the LoRA weight approximation may be compromised due to this discrepancy in rank.
Moreover, even if only a single frequency is suboptimal, the rank of $\mathbf{\widetilde{R}} - \mathbf{I}$
does not become a small number. Taking LLaMA-2 as an example,
each layer of LLaMA-2 contains 32 attention heads.  If there is only one suboptimal frequency, the rank of $\mathbf{\widetilde{R}} - \mathbf{I}$ could reach 32.
In contrast, the LoRA method typically utilizes a low-rank matrix with a rank that does not exceed 16 in practical applications.

Beyond the matter of rank inconsistency, the diversity in the distribution of frequencies, initial phases, and the norms of the embeddings leads to a sophisticated mapping among attention layers, thereby increasing the complexity of the fine-tuning procedure.


\subsection{Phase Shift Calibration (PSC)}
Drawing inspiration from the ResNet \citep{he2016residual} in the field of computer vision, we propose a phase shift calibration module to tackle this issue. Figure \ref{framework} demonstrates the key components of our approach.
We posit that the embedding can be divided into two components:
one is the base embedding, which LoRA can effectively learn;
and the other is shift embedding, which should be acquired separately.
This shift embedding arises from the phase shift discussed in the preceding section.
To be specific, $f^*_q(\mathbf{x}_m, m) \simeq \hat{f}_q\left(\mathbf{P}(\mathbf{x}_m)\odot\mathbf{x}_m+\mathbf{x}_m, m\right)$, where $\mathbf{P}$ presents a two-layer Multilayer Perceptron (MLP) composed of a learnable block diagonal matrix
and $\odot$ is the element-wise production.

In practice, since the frequencies of RoPE are organized block-wise instead of pair-wise \citep{wolf-etal-2020-transformers}, we hence design a head-wise block diagonal matrix.
More specifically:
\begin{equation}
    \mathbf{P}(\mathbf{x}) = \sigma_2\left(\mathbf{W}_2 \left(\sigma_1\left(\mathbf{W}_1 \mathbf{x} \right)\right)\right),
\end{equation}
where $\mathbf{W}_1$ and $\mathbf{W}_2$ are block diagonal matrices with each block size $R^{d_h\times d_h}$, and $d_h$ is the size of single head dimension. For LLaMA and Mistral model, $d_h = 128$, our approach incorporates only a small set of parameters ($<1\%)$, therefore it is parameter efficient.
$\sigma_1$ and $\sigma_2$ are activation functions, we set $\sigma_1$ = SiLU and $\sigma_2 = \frac{1}{2}$Tanh.

There could be two forms of phase shift calibration according to its position:
(1) pre-calibration with the form $\hat{f}_q(\mathbf{P}(\mathbf{x}_m)\odot\mathbf{x}_m + \mathbf{x}_m, m)$ applies phase shift calibration before the position encoding module;
(2) post-calibration which form is $\left(\mathbf{P}\left(\hat{f}_q(\mathbf{x}_m, m)\right)+1\right)\odot\hat{f}_q(\mathbf{x}_m, m)$ applies phase shift calibration after the position encoding module. In the experimental section, we will compare the two forms, and the results show that the positioning of the calibration mechanism affects performance distinctly.
Additionally, our approach is remarkably straightforward to implement.
Algorithm \ref{algorithm} shows the Pytorch-like style of our method.

\begin{algorithm}[t]
  \caption{Pseudocode of phase shift calibration in Pytorch-like style.}\label{algorithm}
  \textcolor{myblue}{\# $q, k, v$: queries, keys, and values;} \\
  \textcolor{myblue}{\# $\mathbf{W}^{q}_1, \mathbf{W}^{q}_2, \mathbf{W}^{k}_1, \mathbf{W}^{k}_2$: block diagonal matrices reshaped into shape (number heads/number key value heads, head dim, head dim);} \\
$q_t=\text{silu} \left(\text{einsum}(\text{`bnsd,ndr->bnsr'}, q, \mathbf{W}^{q}_1)  \right)$ \\
$p_q=\frac{1}{2}\text{tanh} \left(\text{einsum}(\text{`bnsr,nrd->bnsd'}, q_t, \mathbf{W}^{q}_2) \right)$ \\
$k_t=\text{silu} \left(\text{einsum}(\text{`bnsd,ndr->bnsr'}, k, \mathbf{W}^{k}_1)  \right)$ \\
$p_k=\frac{1}{2}\text{tanh} \left(\text{einsum}(\text{`bnsr,nrd->bnsd'}, k_t, \mathbf{W}^{k}_2) \right)$ \\
$q,k=\text{apply\_rotary\_pos\_emb}(q+p_q*q, k+p_k*k)$\\
$\text{out} = \text{self\_attn}(q,k,v) $ 
\end{algorithm}
\begin{table*}[t]
  \begin{center}
    \begin{tabular}{cc ccc ccc cc}
      \hline
     Extention    & Context & \multicolumn{8}{c}{Evaluation Context Length} \\
     Method       & Window  &  2048 & 4096 &  6144 & 8192 & 10240 & 12288 & 14336 & 16384  \\
  \hline
    -       &  4k      & 8.08      &  7.71    &  39.21    &$>10^2$&  $>10^2$   & $>10^2$ & $>10^2$ &$>10^3$  \\    
   PI       &  16k      &  16.74     &  15.55    &  15.04    & 14.76 &  14.60   & 14.53 &  14.51 & 14.59 \\  
    YaRN       &  16k      & 8.45      &  8.09    &  7.97   & 7.92 &  7.90   & 7.91 & 7.93& 9.44\\  
      \hline
    $\text{PI}_\text{FT}$           &   16k    & 8.20      &  7.79   & 7.61& 7.51 & 7.44     & 7.39& 7.35& 7.32 \\
   $\text{PI}^\text{PSC}_\text{FT}$    &    16k     & \textbf{8.16} & \textbf{7.76} &\textbf{7.58} &\textbf{7.48}& \textbf{7.41} &\textbf{7.36}&\textbf{7.32} & \textbf{7.28} \\
 \hline
    $\text{LongRoPE}_\text{FT}$           &   16k    & 8.04    & 7.68  & 7.52& 7.42 &  7.36& 7.31& 7.28&  7.26 \\
    $\text{LongRoPE}^\text{PSC}_\text{FT}$    &    16k     & \textbf{8.03}& \textbf{7.67} & \textbf{7.51}& \textbf{7.41}& \textbf{7.35} & \textbf{7.30} & \textbf{7.26}&  \textbf{7.24}\\  
      \hline
   $\text{YaRN}_\text{FT}$        &   16k      &  8.07     &  7.70   & 7.53&   7.44  &  7.38   & 7.33 & 7.29 & 7.27\\
   $\text{YaRN}^\text{PSC}_\text{FT}$  &   16k      &  \textbf{8.05}  & \textbf{7.67}  & \textbf{7.51} & \textbf{7.41} & \textbf{7.35} & \textbf{7.30} & \textbf{7.26} & \textbf{7.24}\\
  \hline
    \end{tabular}
  \end{center}
  \caption{Sliding window perplexity (S=256) of ten 96k PG19 documents over LLaMA-2 7B. The ``-'' means the base LLaMA2 model. $\diamondsuit_\textit{FT}$ means the extended model is fine-tuned with LoRA (r=8).  $\diamondsuit^\textit{PSC}_\textit{FT}$ means the extended model is fine-tuned with LoRA (r=8) and injected with the PSC module.}
  \label{tbl:16k:ppl}  
\end{table*}
  
\begin{table*}[t]
  \begin{center}
    \begin{tabular}{ccc ccc ccc ccc}
      \hline
     Extention    & Context & \multicolumn{8}{c}{Evaluation Context Length} \\
     Method       & Window  &  4096 & 8192 &  12288 & 16384 & 20480 & 24576 & 28672 & 32768 \\
  \hline
    $\text{PI}_\text{FT}$           &   32k    &  7.95     &  7.65 & 7.53&  7.44   & 7.39 & 7.36 & 7.34 &7.34\\
   $\text{PI}^\text{PSC}_\text{FT}$    &    32k     &  \textbf{7.88}   & \textbf{7.60}  & \textbf{7.47}& \textbf{7.38}&  \textbf{7.33}  & \textbf{7.30}& \textbf{7.28}& \textbf{7.27}\\
      \hline
    $\text{YaRN}_\text{FT}$        &   32k      &   7.76   & 7.49    &7.38  & 7.31  & 7.26  & 7.23 & 7.22 & 7.23\\
    $\text{YaRN}^\text{PSC}_\text{FT}$  &   32k      &   \textbf{7.70}   & \textbf{7.44} &  \textbf{7.33}& \textbf{7.26} & \textbf{7.21} & \textbf{7.19} & \textbf{7.17} & \textbf{7.17}\\
      \hline
    \end{tabular}
  \end{center}
  \caption{Sliding window perplexity (S=256) of ten 96k PG19 documents over LLaMA-2 7B (32k).}
  \label{tbl:32k:ppl} 
\end{table*}
\begin{table*}[t]
  \begin{center}
    \begin{tabular}{ccc ccc ccc ccc}
      \hline
     Extention    & Context & \multicolumn{8}{c}{Evaluation Context Length} \\
     Method       & Window  &  4096 & 8192 & 16384 & 24576 & 32768 & 40960 & 49152 & 57344 &65536 \\
     \hline
    $\text{PI}_\text{FT}$           &   64k    &  8.18   & 7.87 & 7.65&  7.57 &  7.53&7.52 & 7.51 & 7.49 & 7.48\\
   $\text{PI}^\text{PSC}_\text{FT}$  &  64k     & \textbf{8.09}  & \textbf{7.79} & \textbf{7.57} & \textbf{7.49}& \textbf{7.46} & \textbf{7.44} & \textbf{7.43} & \textbf{7.41} & \textbf{7.39}\\
     \hline
    $\text{YaRN}_\text{FT}$           &   64k    &  7.85  & 7.59 & 7.41& 7.34 &7.32 & 7.32& 7.30&7.29 & 7.32\\
   $\text{YaRN}^\text{PSC}_\text{FT}$  &  64k     & \textbf{7.75}  & \textbf{7.49} & \textbf{7.31} & \textbf{7.25}& \textbf{7.23} & \textbf{7.22} & \textbf{7.21} & \textbf{7.19} & \textbf{7.19}\\
      \hline      
    \end{tabular}
  \end{center}
  \caption{Sliding window perplexity (S=256) of ten 96k PG19 documents over LLaMA-2 7B (64k).}
  \label{tbl:7b-64k:ppl}  
\end{table*}

\section{Experiments}
We demonstrate that phase shift calibration successfully realizes the context window extension of large language models by using RoPE extensions as its position encoding schemes. Furthermore, our approach is compatible with a broad range of position encoding techniques, including search-based methods (LongRoPE), position interpolation (PI), and the combination of interpolation and extrapolation techniques (YaRN).
\subsection{Experimental Settings}
\paragraph{Model.}
We conduct experiments on LLaMA-2 and Mistral with various position encode approaches.
In addition, we assess our approach by utilizing several well-known publicly available models, including Together.ai \citep{together_ai}, CodeLlama \citep{Rozire2023CodeLO}, and LongLoRA \citep{longlora}.
\paragraph{Datasets.}
In order to comprehensively and meticulously analyze our technique, we employ several datasets to train and assess our context-extended model. We initially carry out experiments by utilizing a small dataset sampled from the RedPajama \citep{together_computer} dataset, and the length of each text in the sampled dataset is greater than 4K.
We also utilize the PG19 \citep{Rae2019CompressiveTF} train split dataset chunked into 64k segments for training.
While conducting the evaluation, we use the PG19 validation split and the Proof-pile \citep{proof_pile} test split. Details are shown in the appendix.


\subsection{Evaluation}
\paragraph{Long-sequence Language Modeling.}
We make a comparison of the long sequence language modeling performance using the perplexity metric. The sliding window method from \citep{alibi2022} with S=256 is adopted.

We initially present the evaluation results on the LLaMA-2 model and its context window extensions using various approaches in Table \ref{tbl:16k:ppl}, Table \ref{tbl:32k:ppl}, and Table \ref{tbl:7b-64k:ppl}. We extend LLaMA-2 with diverse position encoding schemes such as PI, YaRN, and LongRoPE.
When fine-tuning, we employ LoRA with a rank of 8. We can notice that the fine-tuned models show lower perplexity than the non-fine-tuned ones. Phase shift calibration can enhance all the base position encoding schemes. It even boosts the performance of the optimal-based method LongRoPE. The possible reason might be that the scale factor search space is exponential, which makes it difficult to search for an ideal frequency, and the objective signal may be too sparse as only 5 PG19 texts are used to guide the search. More significantly, by comparing Table \ref{tbl:16k:ppl}, Table \ref{tbl:32k:ppl}, and Table \ref{tbl:7b-64k:ppl}, we can notice that the advantage of applying phase shift calibration becomes greater as the extended context window changes from 16k to 64k. The reason perhaps is that as the context window increases, the largest possible rescale factor also increases. In other words, the frequency solution spaces are enlarged, which makes it even more difficult to predefine an ideal frequency. With phase shift calibration, the frequencies are pre-calibrated to an ideal position.

\begin{table*}[t]
  \begin{center}
    \begin{tabular}{ccc ccc cc c}
      \hline
  Model &  Model &   Extention    & Context & \multicolumn{5}{c}{Evaluation Context Length} \\
  Size &  Name  &    Method       & Window  &  4096 & 8192 &  16384 & 32768  & 65536 \\
      \hline 
  7B &     Together         &   PI    &  32k & 2.47& 2.31&  2.19& 2.11&  $>10^2$\\
  7B &     $\text{Together}_{\text{PSC}}$  &    PI     & 32k & \textbf{2.46}  & \textbf{2.30} & \textbf{2.18} & \textbf{2.10}& $>10^2$\\
      \hline   
  7B &     CodeLlama         &   NTK    &  100k & 2.57& 2.38 &  2.25& 2.16&  2.15\\
  7B &     $\text{CodeLlama}_{\text{PSC}}$  &    NTK     & 100k & 2.57  & 2.38&  \textbf{2.24} & \textbf{2.15}& \textbf{2.12}\\
      \hline   
  7B &     LongLoRA         &   PI    &  32k & 2.57& 2.38& 2.25& 2.16& $>10^2$\\
  7B &     $\text{LongLoRA}_{\text{PSC}}$  &   PI     & 32k & \textbf{2.50}  & \textbf{2.32}& \textbf{2.20}& \textbf{2.12}& $>10^2$\\
  \hline
  7B &     YaRN          &   YaRN   &  64k & 2.50& 2.34 & 2.23& 2.14& 2.08\\
  7B &     $\text{YaRN}_{\text{PSC}}$  &   YaRN     & 64k&\textbf{2.46} & \textbf{2.30}& \textbf{2.19}& \textbf{2.11} &  \textbf{2.05}\\
      \hline  
    \end{tabular}
  \end{center}
  \caption{Sliding window perplexity (S=256) of ten 128k Proof-pile documents over various models.}
  \label{tbl:ppl:pub}  
\end{table*}

We also incorporate the phase shift calibration module into several well-known publicly available models, like Together.ai, CodeLlama, and LongLoRA. We fine-tune the enhanced model using the PG19 dataset and assess it on the Proof-pile dataset. Table \ref{tbl:ppl:pub} presents the outcomes, and we can note that phase shift calibration enhances LongLora and YaRN more prominently than it does for Together and CodeLlama. This may be due to that the Together and CodeLlama are pre-trained and fine-tuned with full parameter updates, while the remaining ones utilize the LoRA-like method.

\paragraph{Passkey Retrieval.}
\begin{figure}[t]
\begin{center}
  \includegraphics[width=\linewidth]{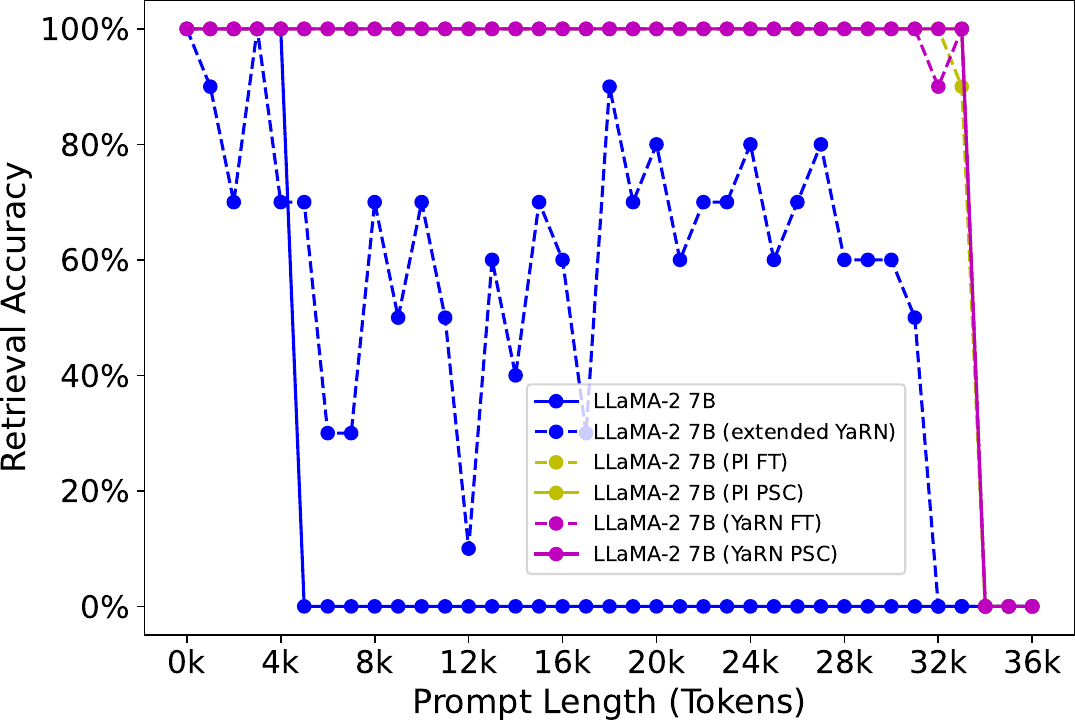}
\end{center}
\caption{A comparison of passkey retrieval accuracy for context-augmented Large Language Models (LLMs).
"Extend YaRN" indicates that the model incorporates YaRN without undergoing fine-tuning. "FT" denotes that the models have been fine-tuned using LoRA (r=8), while "PSC" signifies that the models have been fine-tuned with the phase shift calibration module activated. (The graphs for LLaMA-2 7B (PI PSC) and LLaMA-2 (YaRN PSC) coincide as they exhibit the same results: with 100\% accuracy up to 34k.)}
\label{fig:pass_acc}
\end{figure}

The passkey retrieval task proposed by \citep{Mohtashami2023LandmarkAR} gauges a model's effective context window size. This task aims to require a model to fetch a simple passkey from a large set of useless tokens. In our assessment, we conduct 10 iterations of the passkey retrieval task with the context window sizes ranging from 2k to 36k. The random passkey is positioned at a random location that is uniformly distributed among the collection of the tokens. The prompt template is presented in the appendix.

The comparison of retrieval accuracy with various approaches is presented in Figure \ref{fig:pass_acc}. We can notice that the accuracy of the LLaMA-2 base model drops instantly to 0 when the sequence length goes beyond its pre-trained context window length. Although extending the context window using YaRN without fine-tuning can raise the accuracy beyond the 4k pre-trained context size, the accuracy is lower and the performance is less stable compared to the fine-tuning-based models. With fine-tuning, position encoding methods such as PI and YaRN can significantly enhance the retrieval accuracy. However, the accuracy becomes unstable as the evaluated context length gets closer to the context window size. For example, at 32k, the accuracy of LLaMA-2 7B (YaRN FT) drops to 90\%, while at 33k, the accuracy of LLaMA-2 7B (PI FT) drops to 90\%. Both LLaMA-2 7B (PI PSC) and LLaMA-2 7B (YaRN PSC) show a 100\% retrieval accuracy up to a 34k context length when the phase shift calibration module is enabled.


\paragraph{Standard Benchmarks}
\begin{table*}[t]
  \begin{center}
    \begin{tabular}{cccc cccc}
      \hline
  Model &  Model &   Extention    & Context & \multirow{2}{*}{ARC-c} & \multirow{2}{*}{HellaSwag} & \multirow{2}{*}{MMLU} & \multirow{2}{*}{TruthfulQA} \\
  Size &  Name  &    Method       & Window  &   & &  &    \\
  \hline
  7B &  Llama2 &       -          &  4k &    \underline{52.47}   & \textbf{78.97}& \textbf{46.24} & \underline{38.96} \\

  7B &     Together         &   PI    &  32k & 47.27& 77.41& 45.33 & 38.4\\
  7B &     $\text{Together}_{\text{PSC}}$  &    PI     & 32k & 47.35  & 77.39& \underline{45.57}& 37.66\\

  7B &     CodeLlama         &   NTK    &  100k & 43.69& 65.03&39.56   & 37.2\\
  7B &     $\text{CodeLlama}_{\text{PSC}}$  &    NTK     & 100k & 42.75 & 64.81&39.77 & 36.31\\

  7B &     LongLora         &   PI    &  32k & 50.51& 76.32& 37.81& 37.92\\
  7B &     $\text{LongLora}_{\text{PSC}}$  &   PI     & 32k & 50.60& 76.82& 39.39& 38.71\\

  7B &     YaRN         &   YaRN   &  64k & \textbf{52.99}&\underline{78.25} & 42.46& 38.32\\
  7B &     $\text{YaRN}_\text{PSC}$  &   YaRN     & 64k& 52.30& 78.11& 42.12 & \textbf{39.81} \\
      \hline  
    \end{tabular}
  \end{center}
  \caption{Performance of context-extended methods on the Hugging Face Open LLM benchmark suite. }
  \label{tbl:openllm}
\end{table*}
We assess different methods in comparison with the original LLaMA-2 model by using the Hugging Face Open LLM Leaderboard \citep{open_llm_leaderboard}. Specifically, the Language Model Evaluation Harness library \citep{eval-harness} is utilized to carry out the evaluation. We employ 25-shot ARC-Challenge \citep{Clark2018ThinkYH}, 10-shot HellaSwag \citep{Zellers2019HellaSwagCA}, 5-shot MMLU \citep{Hendrycks2020MeasuringMM}, and 0-shot TruthfulQA \citep{Lin2021TruthfulQAMH}.

The experiments aim to assess the degradation of model performance along with the context-extended window.
We compare different models equipped with the phase shift calibration module with the relevant baselines and the original LLaMA-2 model. The results are summarized in Table \ref{tbl:openllm}. We can notice that models armed with the phase shift calibration show comparable performance to the related baselines. PSC can even outperform the related baselines. For instance, $\text{LongLoRA}_{\text{PSC}}$ outperforms  LongLoRA on all datasets, $\text{Together}_{\text{PSC}}$  attains the second-best performance on the MMLU dataset.
Even more notable,  $\text{YaRN}_\text{PSC}$ even achieves the best performance on the TruthfulQA dataset, with the accuracy performance increased by 0.85\%.


\begin{table*}[t]
  \begin{center}
    \begin{tabular}{ccc ccc ccc cc}
      \hline
  \multirow{2}{*}{Rank} &     Extention    & Context & \multicolumn{8}{c}{Evaluation Context Length} \\
  &     Method       & Window  &  4096 & 8192 &  12288 & 16384 & 20480 & 24576 & 28672 & 32768  \\
      \hline
  12 &    $\text{YaRN}_\text{FT}$           &   32k    &  7.76  & 7.49 & 7.38 & 7.30 & 7.26 & 7.23 & 7.22 & 7.22\\
  12 &   $\text{YaRN}^\text{PSC}_\text{FT}$    &    32k     & \textbf{7.70} & \textbf{7.43} & \textbf{7.33}& \textbf{7.26}&\textbf{7.21} & \textbf{7.18} &\textbf{7.17}& \textbf{7.17}\\
      \hline
 16 &    $\text{YaRN}_\text{FT}$           &   32k    &  7.75  &  7.49& 7.37 & 7.30 &  7.26& 7.23 & 7.22& 7.23\\
 16 &   $\text{YaRN}^\text{PSC}_\text{FT}$    &    32k     & \textbf{7.70} & \textbf{7.44} & \textbf{7.33}& \textbf{7.26}& \textbf{7.21}& \textbf{7.18}& \textbf{7.18} & \textbf{7.17}\\
      \hline
    \end{tabular}
  \end{center}
  \caption{Sliding window perplexity (S=256) of ten 96k PG19 documents over LLaMA-2 7B.}
  \label{tbl:rank:ppl}  
\end{table*}
\begin{table*}[t!]
  \begin{center}
    \begin{tabular}{cc ccc ccc cc}
      \hline
     Extention    & Context & \multicolumn{8}{c}{Evaluation Context Length} \\
     Method       & Window  &  4096 & 8192 &  12288 & 16384 & 20480 & 24576 & 28672 & 32768  \\
      \hline
    $\text{PI}^\text{PSC}_\text{before}$           &   32k    &  \textbf{8.24}  & \textbf{7.93} & \textbf{7.80 }& \textbf{7.72}& \textbf{7.66} &\textbf{7.63}& \textbf{7.62}& \textbf{7.62}\\
    $\text{PI}^\text{PSC}_\text{after}$    &    32k     & 8.50 & 8.21 & 8.10& 8.05& 8.02& 8.03&8.07 & 8.16\\
      \hline
    \end{tabular}
  \end{center}
  \caption{Sliding window perplexity (S=256) of ten 96k PG19 documents over LLaMA-2 7B.}
  \label{tbl:psc-only}  
\end{table*}
\paragraph{Long Context Benchmarks}

\begin{table*}[t!]
  \begin{center}
    \begin{tabular}{ccc ccc ccc ccc}
    \hline
    Model      & Tokens & Coursera &  GSM & QaALITY & TOFEL & CodeU & SFiction &  Avg. \\
      \hline
Llama2-7B       & 4k     &    15.26 & 19.0 &   30.69 & 13.01 & 3.33  &    35.93 & 19.54 \\
      \hline
 $\text{PI}_\text{FT}$    & 16k    &    16.86 & 18.0 &   27.23 & 33.45 &  3.33 &    39.06 & 22.99 \\
$\text{PI}^\text{PSC}_\text{FT}$      & 16k    & \textbf{20.64} & 18.0 &  \textbf{29.70} & 30.48 &  3.33 &    \textbf{43.75} & \textbf{24.32} \\
      \hline
 $\text{PI}_\text{FT}$    & 64k    &    21.07 & 11.0 &   13.37 & 22.30 &  1.11 &    40.62 & 18.25 \\
 $\text{PI}^\text{PSC}_\text{FT}$    & 64k    &    20.21 & \textbf{14.0} &   \textbf{26.24} & \textbf{25.28} &  3.33 &    \textbf{42.19} & \textbf{21.88} \\
\hline
    \end{tabular}
  \end{center}
  \caption{Evaluation results on L-Eval benchmarks.}
  \label{tbl:leval}
\end{table*}
We also evaluated our method using the L-Eval benchmarks \citep{an2023leval}. L-Eval is a comprehensive evaluation suite designed to assess long-context language models across multiple sub-tasks. Our experiments were performed on the Llama2-7B model, utilizing the PI method both with and without PSC. The results are detailed in Table \ref{tbl:leval}.
From these results, we observe that enabling PSC contributes to an improvement in the average L-Eval score at both 16k and 64k contexts. However, the average score at 64k is noted to be lower than at 16k. This discrepancy may be attributed to the increase in perplexity as the context window expands. Additionally, the average length of many datasets is shorter than 16k, which could influence the performance at 64k.
\subsection{Ablation Study}
In this section, we present ablation studies on the phase shift calibration modules. We aim to address the following questions:
(1) Since the phase shift calibration module introduces a few additional parameters, can a LoRA with a large rank outperform the PSC module?
(2) What is the effectiveness of the phase shift calibration module at different positions of the base model?
(3) What is the performance of the phase shift calibration with respect to the number of fine-tuning steps?
\paragraph{More Parameters.}
We fine-tune the base model with different ranks and position encoding methods and assess the performances. The results are presented in Table \ref{tbl:rank:ppl}. Several discoveries are apparent. First, with phase shift calibration, we can obtain stable improvements across various token lengths at different ranks. Second, increasing the rank size of LoRA leads to almost no performance gain. Additionally, Table \ref{tbl:32k:ppl} shows the results of model fine-tuning with LoRA rank 8. By comparing it with Table \ref{tbl:rank:ppl}, we can observe that even if the LoRA rank is doubled, the performance gains are negligible. Hence, the performance of phase shift calibration does not stem from more parameters but from calibrating the frequencies to the optimal states.
\paragraph{Pre-calibration vs Post-calibration.}
We evaluate the effectiveness of the phase shift calibration module at different positions. In this experiment, we only update the parameters of the PSC while keeping the other parameters frozen. The results are summarized in Table \ref{tbl:psc-only}. We have several key findings with these results. First, by comparing it with Table \ref{tbl:32k:ppl}, we can observe that the phase shift calibration itself can improve the perplexity of the models. When combined with LoRA, it can further enhance the performance. Second, applying the phase shift calibration before the position encoding method is better than applying it after the position encoding method. The possible reason is that the position encoding method introduces complex non-linear distortion to the query/key embeddings.

\paragraph{Ablation on Fine-tuning Steps.}
\begin{figure}[t]
\begin{center}
  \includegraphics[width=\linewidth]{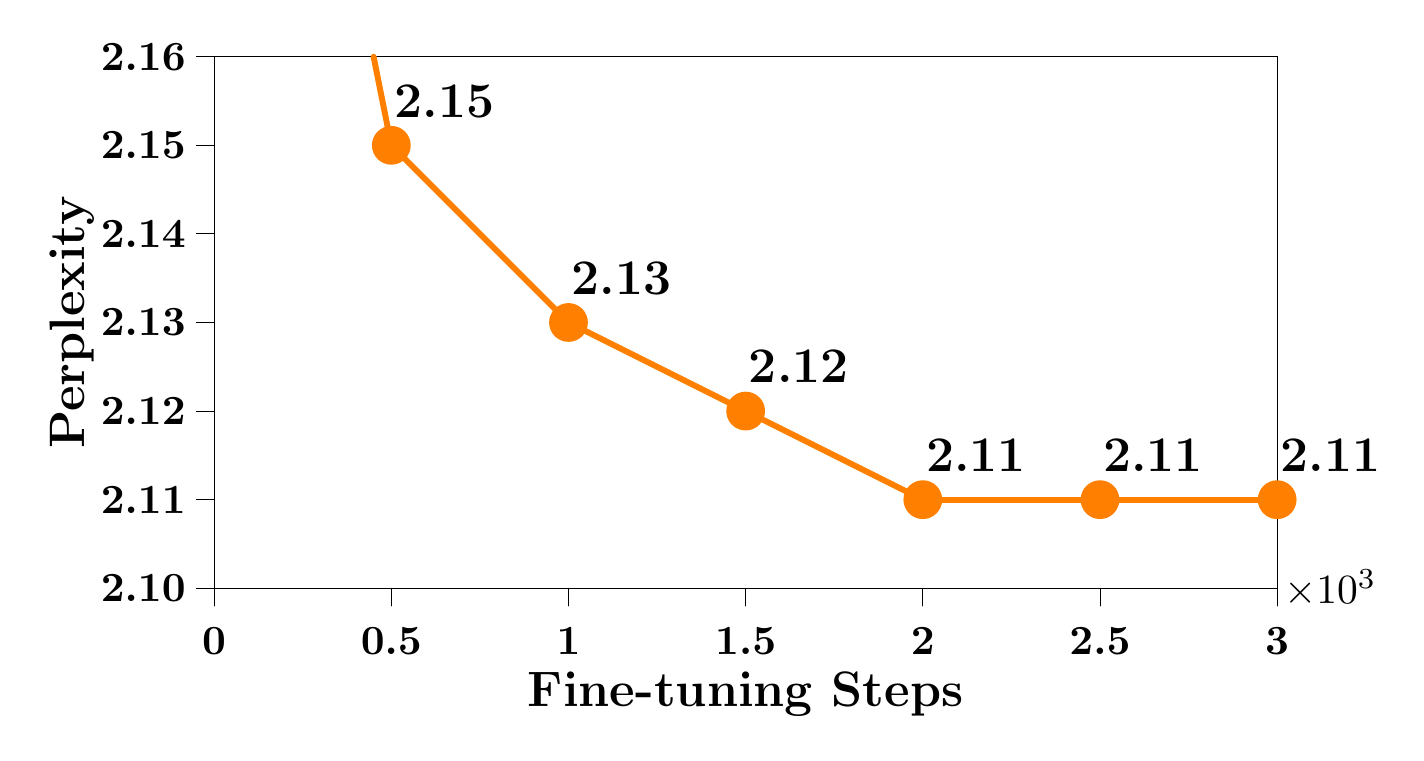}
\end{center}
\caption{An ablation study on the fine-tuning process utilizing phase shift calibration. The perplexity is assessed with a context length of 32k.}
\label{fig:mistral_steps}
\end{figure}
We present the relationship between perplexity and fine-tuning steps for the Mistral-7B model extended to a 32K context window on the Proof-pile test set. As Figure \ref{fig:mistral_steps} indicates, the perplexity drops rapidly to 2.15 at step 500, and then gradually converges to 2.11 at step 2000. Further fine-tuning the model from step 2000 does not lead to any further improvement.
Thus, a stopping criterion can be implemented to conserve computational resources. Calculating the perplexity for the entire dataset is computationally expensive. Instead, we might opt to sample a subset of documents to approximate the perplexity, using this estimation as our stopping criterion. Additionally, setting a baseline number of steps and applying the stopping criterion only after surpassing this baseline can further alleviate the computational burden linked with perplexity calculations. For models tailored to specific domains, employing domain-specific metrics as stopping criteria can be a judicious approach, offering a more precise evaluation of the model’s effectiveness within that particular context.

\subsection{Complexity}
\begin{figure}[t]
\begin{center}
  \includegraphics[width=\linewidth]{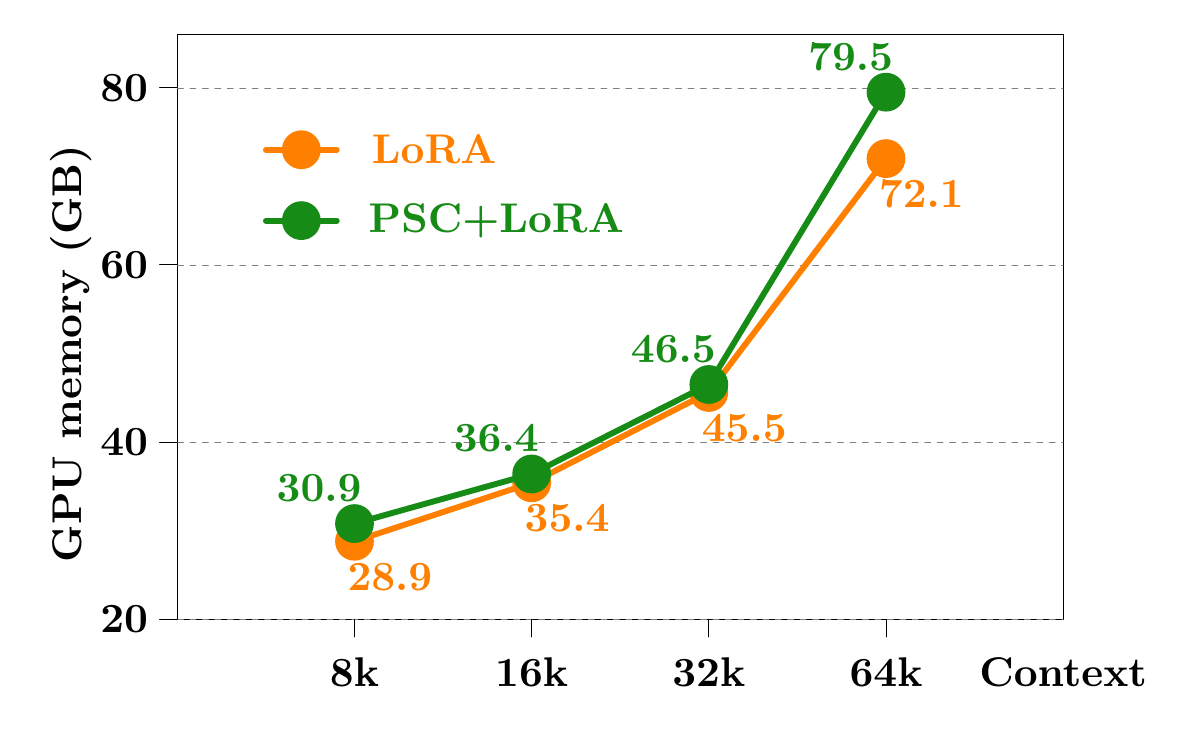}
\end{center}
\caption{GPU memory consumption by LoRA and PSC.}
\label{fig:gpu_mem}
\end{figure}
\begin{table}[!t]
  \begin{center}
    \begin{tabular}{c c c c c}
        \hline
      Method & Window & Tokens & time(ms)  \\
      \hline
   LoRA  & 32k            & 16k    &   1686.0          \\
  PSC+LoRA & 32k          & 16k    &   1691.6        \\
      \hline      
    \end{tabular}
  \end{center}
  \caption{Computational overhead with the PI position encoding.}
  \label{tbl:inf:overhead}    
\end{table}

Phase shift calibration defines block diagonal matrices for query/key embeddings.
Each block is a $d_h \times d_h$ matrix, where $d_h$ is the dimension of a single head.
As a result,  it introduces additional 64M parameters for LLaMA-2 7B, accounting for 0.095\% ($< 1\%)$ of the total parameters. Figure \ref{fig:gpu_mem} shows the GPU memory used by LoRA and PSC. To assess the computational overhead, we extend the LLama2-7B model with the PI method and perform the next token prediction task with batch size 1. The results are presented in Table \ref{tbl:inf:overhead}, where the \textit{Tokens} means the number of input tokens we feed into the model.

\section{Conclusion}
In this work, we present PSC: Phase Shift Calibration, an approach for calibrating the existing extended position encoding methods.
We first present that there is a rank inconsistency issue when the pre-defined frequencies are not optimal.
A phase shift calibration module is designed to remedy this issue.
We conduct extensive experiments on various tasks, and the results show that PSC is compatible with various context extension methods, including interpolation, mixing of interpolation/extrapolation, and search-based techniques. With PSC, the long-range abilities of LLMs can be further enhanced. Moreover, our method only introduces a few more parameters ($<1\%$), which is parameter-efficient. This work thus supports many natural language processing tasks that require long-range capabilities. We discuss several promising future works in the appendix.

\section{Limitations}
This paper introduces a phase shift calibration module to the base model to further enhance the performance of existing position encoding methods.
Since the introduced phase shift calibration module contains a small set of trainable parameters, our method requires fine-tuning of the enhanced models and needs a bit more GPU memory than simply fine-tuning with LoRA.

\bibliography{emnlp_2024}

\appendix

\newpage
\section{Appendix}
\label{sec:appendix}
\subsection{Settings}
\paragraph{Training.}
For training, we employ the AdamW optimizer \citep{Loshchilov2019DecoupledWD} with $\beta_1 = 0.9$ and $\beta_2 = 0.95$. We utilize a learning rate of $2 \times 10^{-4}$ when training on the sampled RedPajama dataset, and $2 \times 10^{-5}$ otherwise. The weight decay is set to zero, and a linear warmup of 20 steps is applied. All experiments are conducted using the Transformers \citep{wolf-etal-2020-transformers} framework, and Flash Attention 2 \citep{dao2022flashattention, dao2023flashattention2} is utilized to optimize memory usage. For a fair comparison, all models are trained for 3000 steps on 4 A800 GPUs. We set the batch size to the value that maximizes GPU memory utilization and adopt a gradient accumulation step size of 4.
When training LongRoPE, we add three additional rescale factors corresponding to PI, NTK, and YaRN to the initial population. 

\paragraph{Evaluation.}
When training our model with the RedPajama dataset, we evaluate our method by using the PG19 validation split. We pick 10 random samples from the PG19 validation split with at least 96k tokens. When we train our model on the PG19 train split dataset chunked into 64k segments, we evaluate the model using the Proof-pile \citep{proof_pile} test split. Likewise, we select 10 random samples from Proof-pile with at least 128k tokens.

\paragraph{Passkey prompt.}
To measure the effective context window size,
we utilize the prompt employed by  existing literature \citep{Mohtashami2023LandmarkAR, longlora, longrope}.
The prompt is shown as follows:
\begin{tcolorbox}[colback=gray!10,
                  colframe=orange!50,
                  width=\linewidth,
                  arc=2mm, auto outer arc,
                  title={Passkey prompt},breakable,]
There is an important info hidden inside a lot of irrelevant text.
Find it and memorize them.I will quiz you about the important information there.

The grass is green. The sky is blue. The sun is yellow. Here we go.
There and back again. (repeat $M$ times)

The pass key is <PASS KEY>. Remember it. <PASS KEY> is the pass key.

The grass is green. The sky is blue. The sun is yellow. Here we go.
There and back again. (repeat $N$ times)

What is the pass key? The pass key is
\end{tcolorbox}
The <PASS KEY> is the number to retrieve, we randomly generate a passkey in the range $[1, 50000]$ during each testing time.
The text length varies with the values of $M$ and $N$.

\subsection{More Experiments}


\begin{table*}[ht!]
  \begin{center}
    \begin{tabular}{cc ccc ccc cc}
      \hline
     Extention    & Context & \multicolumn{8}{c}{Evaluation Context Length} \\
     Method       & Window  &  4096 & 8192 &  12288 & 16384 & 20480 & 24576 & 28672 & 32768 \\
  \hline
    -      &  8k      & 2.23    & 2.09   & 4.60  & 26.25 &  79.97 & $>10^2$  & $>10^2$&  $>10^2$  \\  
      \hline
    $\text{YaRN}_\text{FT}$        &   32k      & 2.42    & 2.25  & 2.20 & 2.17 & 2.15 & 2.14 & 2.13 & 2.13 \\
    $\text{YaRN}^\text{PSC}_\text{FT}$  &   32k      &  \textbf{2.40}   & \textbf{2.23} & \textbf{2.17} & \textbf{2.15} & \textbf{2.14}& \textbf{2.13}  & \textbf{2.12}& \textbf{2.11} \\
      \hline
    \end{tabular}
  \end{center}
  \caption{Sliding window perplexity (S=256) of ten 128k Proof-pile documents over Mistral 7B. The ``-'' means the base Mistral 7B v0.1 model. $\diamondsuit_\textit{FT}$ means the extended model is fine-tuned with LoRA (r=8).  $\diamondsuit^\textit{PSC}_\textit{FT}$ means the extended model is fine-tuned with LoRA (r=8) and injected with the PSC module.}
  \label{tbl:mistral:32k}  
\end{table*}
\paragraph{Mistral 7B.} We also extend the Mistral 7B v0.1 model \citep{Jiang2023Mistral7},  which is another famous open-source model.
We extend Mistral with YaRN \citep{peng2023yarn} to 32k and perform an ablation study on the phase shift calibration module.
For training, we use a small dataset sampled from the RedPajama \cite{together_computer} dataset with token length $\geq$ 4k. we utilize a constant learning rate $2\times 10^{-4}$ with a linear warmup of 20 steps. We fine-tune the models for 3000 steps.
We evaluate the models using Proof-pile \cite{proof_pile} test split and 10 documents with token length $\geq$ 128k are sampled.
The results are described in Table \ref{tbl:mistral:32k}.
We can observe that with the phase shift calibration module enabled,
the performance of long-range abilities
gets further improved upon YaRN.

\begin{table*}[ht!]
  \begin{center}
    \begin{tabular}{cc ccc ccc cc}
      \hline
       Extention    & Context & \multicolumn{8}{c}{Evaluation Context Length} \\
       Method       & Window  &  2048 & 4096 &  6144 & 8192 & 10240 & 12288 & 14336 & 16384  \\
       \hline
       -       &  4k      &   7.25   & 6.91  & 48.98   &$>10^2$& $>10^3$   & $>10^3$ & $>10^3$ & $>10^3$  \\    
      PI       &  16k      &    12.42  &  11.65   &  11.30   & 11.11 &  10.96  &  10.87& 10.81  & 10.80\\  
      YaRN       &  16k      &  7.54    &  7.22   & 7.10 & 7.06& 7.03  & 7.03 & 7.04& 8.14\\  
      \hline
      $\text{PI}_\text{FT}$           &   16k    &   7.35   & 6.99  & 6.84& 6.75& 6.68 & 6.64 & 6.60& 6.57 \\
      $\text{PI}^\text{PSC}_\text{FT}$    &    16k     & \textbf{7.32} & \textbf{6.97} & \textbf{6.82} &\textbf{6.73}& \textbf{6.67}& \textbf{6.62}& \textbf{6.58}& \textbf{6.55}\\
      \hline
     $\text{YaRN}_\text{FT}$        &   16k      & 7.26   & 6.92   & 6.76&  6.68 & 6.62  & 6.57& 6.54& 6.52\\
     $\text{YaRN}^\text{PSC}_\text{FT}$  &   16k      &\textbf{7.23}   & \textbf{6.89}  &  \textbf{6.74}& \textbf{6.65} & \textbf{6.60} & \textbf{6.55} & \textbf{6.52} &  \textbf{6.49} \\
  \hline
    \end{tabular}
  \end{center}
  \caption{Sliding window perplexity (S=256) of ten 96k PG19 documents over LLaMA-2 13B. The ``-'' means the base LLaMA2 model. $\diamondsuit_\textit{FT}$ means the extended model is fine-tuned with LoRA (r=8).  $\diamondsuit^\textit{PSC}_\textit{FT}$ means the extended model is fine-tuned with LoRA (r=8) and injected with the PSC module.}
  \label{tbl:13b-16k:ppl}  
\end{table*}

\paragraph{LLaMA-2 13B.}
In addition, we assess our approach on the LLaMA-2 13B model \citep{Touvron2023Llama2O}.
The models are fine-tuned with sampled documents from
RedPajama \cite{together_computer} dataset. Each document has token length $\geq $ 4k.
We set the learning rate as $2 \times 10^{-4}$ and use a linear warmup of 20 steps.
Both PI \citep{Chen2023ExtendingCW} and YaRN \citep{peng2023yarn}
are employed in our evaluation.
Table \ref{tbl:13b-16k:ppl} shows the results.
The results exhibit similar performance improvement as the evaluations on the LLaMA-2 7B model.
It demonstrates our method is compatible with various LLMs and position encoding approaches.


\subsection{Initial Phase and Norm distribution}
\begin{figure}[t]
\begin{center}
  \includegraphics[width=\linewidth]{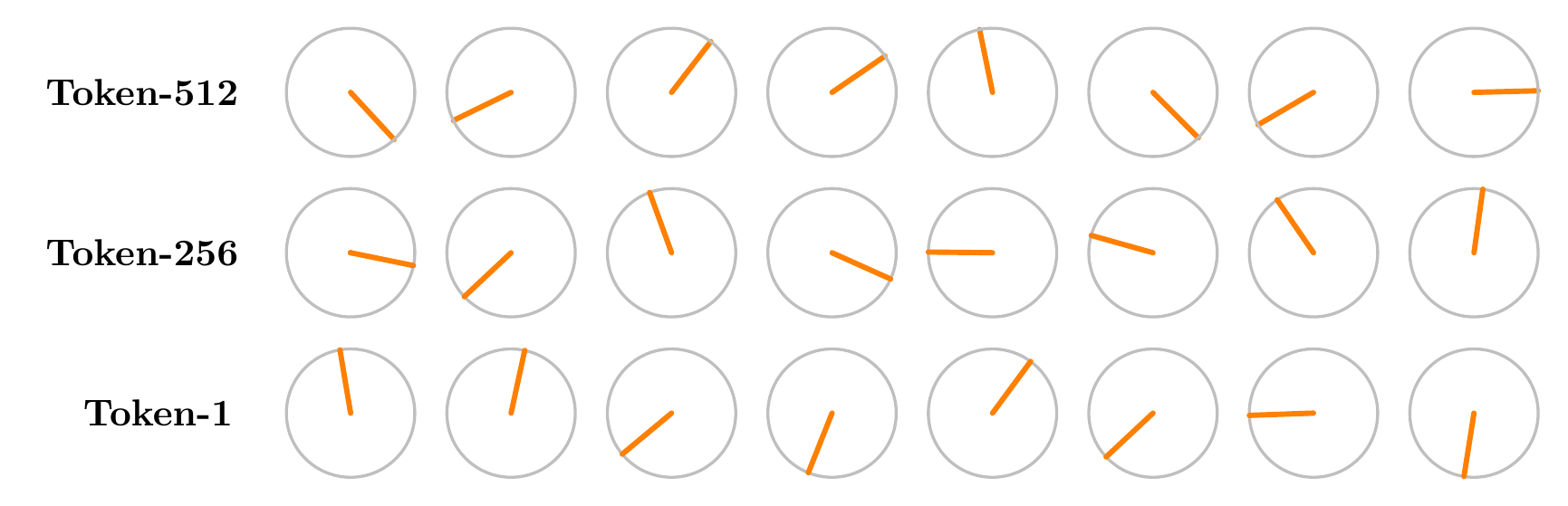}
\end{center}
\caption{The phase of the first eight $(x,y)$ pairs from 3 sampled tokens in layer 6 and head 2 of the LLaMA-2 7B.}
\label{fig:dist:phase:6-2}
\end{figure}

\begin{figure}[t]
\begin{center}
  \includegraphics[width=\linewidth]{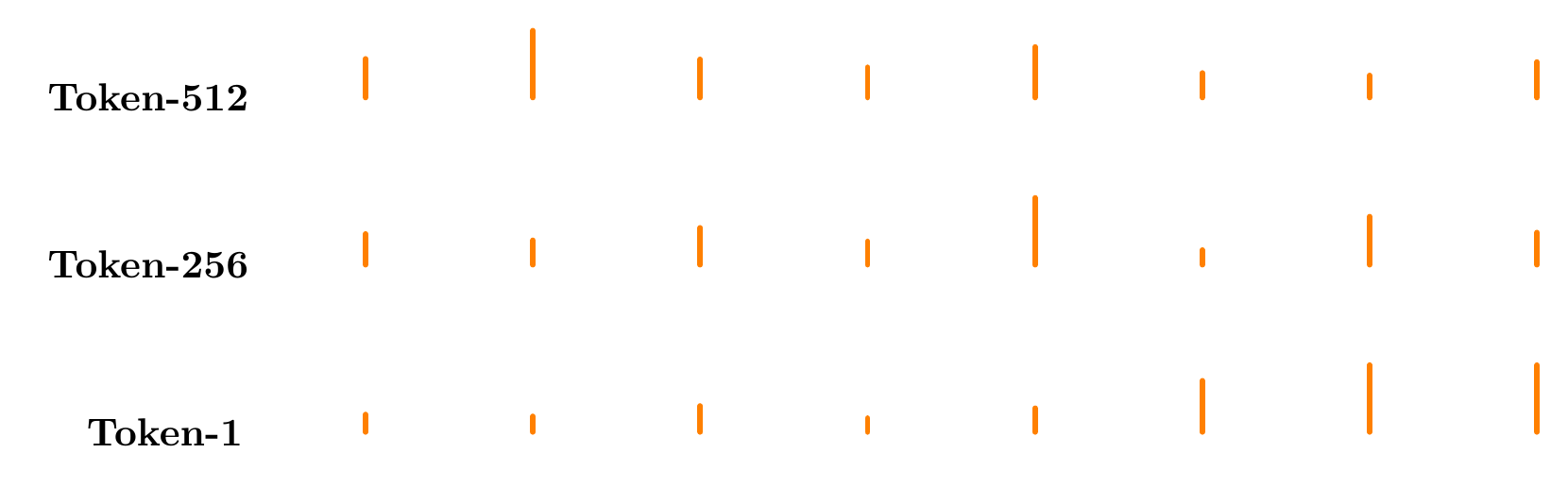}
\end{center}
\caption{The norm of the first eight $(x,y)$ pairs from 3 sampled tokens in layer 6 and head 2 of the LLaMA-2 7B.}
\label{fig:dist:norm:6-2}
\end{figure}

\begin{figure}[t]
\begin{center}
  \includegraphics[width=\linewidth]{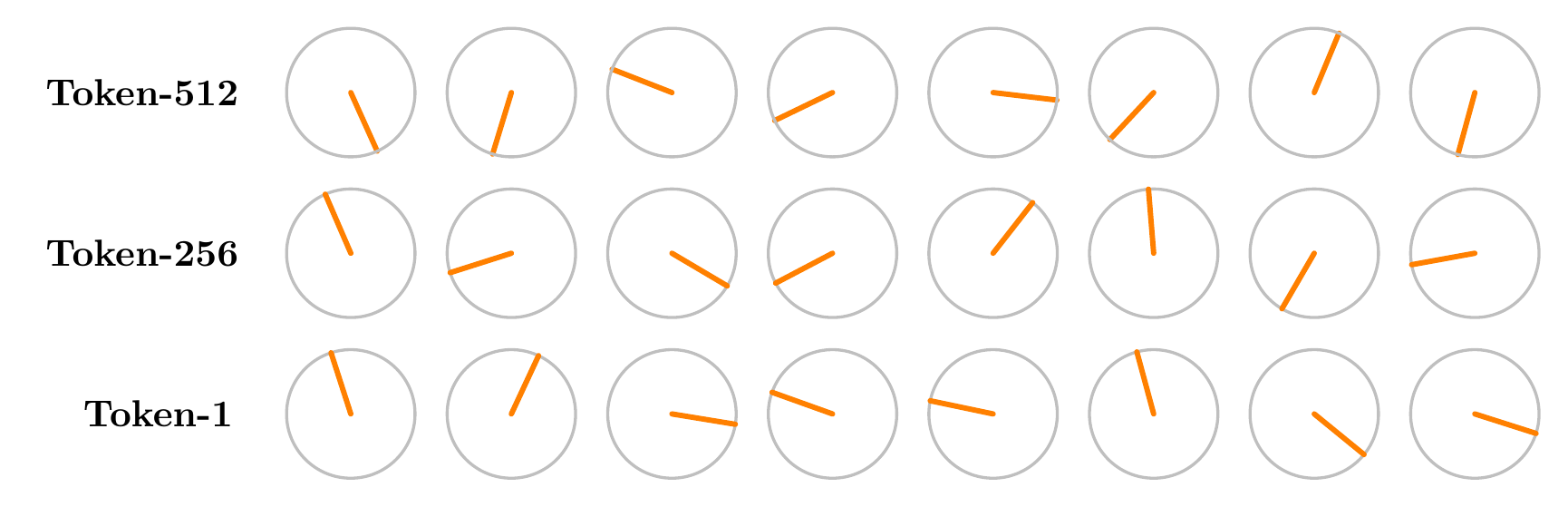}
\end{center}
\caption{The phase of the first eight $(x,y)$ pairs from 3 sampled tokens in layer 6 and head 28 of the LLaMA-2 7B.}
\label{fig:dist:phase:6-28}
\end{figure}

\begin{figure}[t]
\begin{center}
  \includegraphics[width=\linewidth]{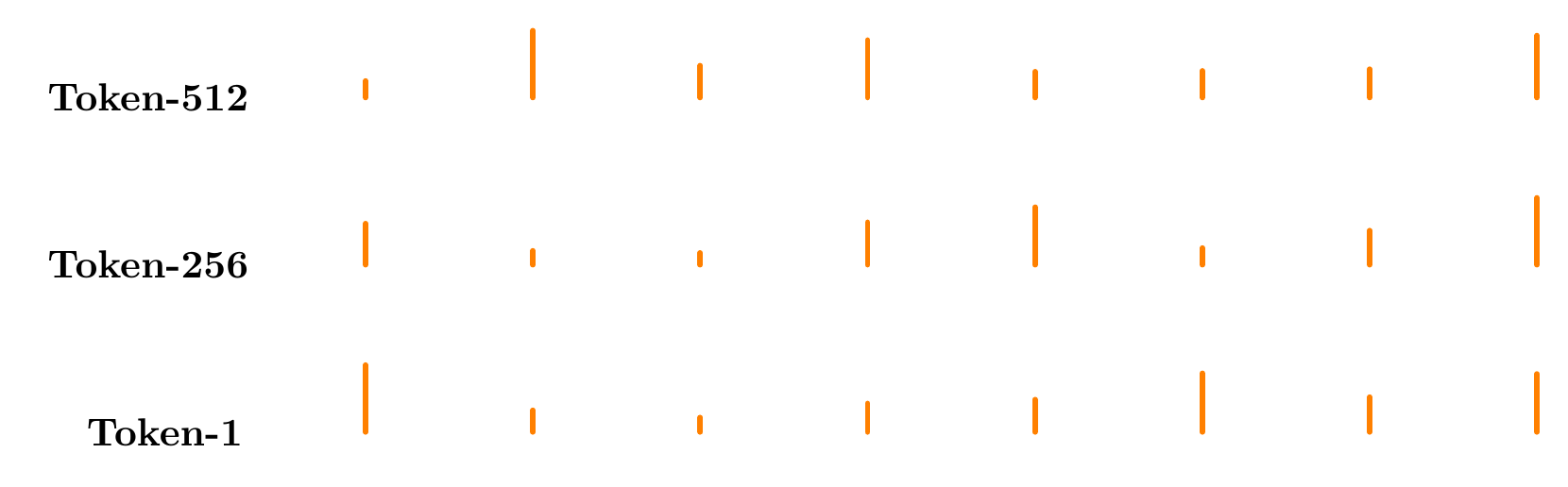}
\end{center}
\caption{The norm of the first eight $(x,y)$ pairs from 3 sampled tokens in layer 6 and head 28 of the LLaMA-2 7B.}
\label{fig:dist:norm:6-28}
\end{figure}

The RoPE and its extensions
consider each pair $(x,y)$ in the embeddings
as a complex number. And perform a rotary transformation
on each pair.
Due the complicated distribution of $(x, y)$,
it is challenging to predefine a set of frequencies
to conduct the rotary transforms.
We show the initial phase and norm distributions of some sampled $(x,y)$
pairs from different layers and heads in
Figure \ref{fig:dist:phase:6-2}
, Figure \ref{fig:dist:norm:6-2}, Figure \ref{fig:dist:phase:6-28},
Figure \ref{fig:dist:norm:6-28},
which have complicated distributions of phase and norm.

\subsection{Future Work}
Our method shows consistent improvements upon various position encoding methods.
For future work, we would investigate PSC applications where long-range
capabilities are needed, such as long-cycle conversations and
LLM-based long-term user historical behavior understanding.
We would also try to seek phase shift calibration methods that
without the need for fine-tuning.

\end{document}